\documentclass[10pt,twocolumn,letterpaper]{article}

\usepackage{wacv}
\usepackage{times}
\usepackage{epsfig}
\usepackage{amsmath}
\usepackage{amssymb}
\usepackage{booktabs}
\usepackage{subcaption}
\usepackage{graphicx}
\usepackage[accsupp]{axessibility} 

\def\wacvPaperID{1509} 

\wacvfinalcopy 

\ifwacvfinal
\usepackage[breaklinks=true,bookmarks=false]{hyperref}
\else
\usepackage[pagebackref=true,breaklinks=true,colorlinks,bookmarks=false]{hyperref}
\fi

\begin{document}

\title{HuPR: A Benchmark for Human Pose Estimation
Using Millimeter Wave Radar}

\author{Shih-Po Lee$^{\dagger}$\thanks{ This work is supported by National Center for High-performance Computing, Taiwan.}, Niraj Prakash Kini$^{\dagger}$, Wen-Hsiao Peng$^{\dagger}$, Ching-Wen Ma$^{\dagger}$, Jenq-Neng Hwang$^{\ddagger}$\\
${}^{\dagger}$National Yang Ming Chiao Tung University, Taiwan\\ ${}^{\ddagger}$University of Washington, USA\\
{\tt\small \{mapl0756051.cs07g,  nirajnctu.cs06g, machingwen\}@nctu.edu.tw, wpeng@cs.nctu.edu.tw, hwang@uw.edu}
}
\maketitle
\thispagestyle{empty}

\begin{abstract}
This paper introduces a novel human pose estimation benchmark, Human Pose with Millimeter Wave Radar (HuPR), that includes synchronized vision and radio signal components. This dataset is created using cross-calibrated mmWave radar sensors and a monocular RGB camera for cross-modality training of radar-based human pose estimation.
There are two advantages of using mmWave radar to perform human pose estimation. First, it is robust to dark and low-light conditions. Second, it is not visually perceivable by humans and thus, can be widely applied to applications with privacy concerns, e.g., surveillance systems in patient rooms.
In addition to the benchmark, we propose a cross-modality training framework that leverages the ground-truth 2D keypoints representing human body joints for training, which are systematically generated from the pre-trained 2D pose estimation network based on a monocular camera input image, avoiding laborious  manual label annotation efforts. The framework consists of a new radar pre-processing method that better extracts the velocity information from radar data, Cross- and Self-Attention Module (CSAM), to fuse multi-scale radar features, and Pose Refinement Graph Convolutional Networks (PRGCN), to refine the predicted keypoint confidence heatmaps. 
Our intensive experiments on the HuPR benchmark show that the proposed scheme achieves better human pose estimation performance with only radar data, as compared to traditional pre-processing solutions and previous radio-frequency-based methods. Our code is available at here\footnote{https://github.com/robert80203/HuPR-A-Benchmark-for-Human-Pose-Estimation-Using-Millimeter-Wave-Radar}
\end{abstract}
\section{Introduction}

\label{sec:intro}
Human pose estimation (HPE) is one of the widely studied traditional tasks in computer vision. Given the RGB images under a single or multiple camera view, it predicts 2D/3D human skeletons, in terms of estimating human body keypoints. Though promising results have been demonstrated by previous HPE solutions, the natural properties of an RGB image undoubtedly constrain the advancement of HPE. In particular, the RGB images captured in the dark and a low-light conditions can hardly show a person's pose, leading to an inferior quality of pose estimation. In addition, using such vision-based inputs consequently raise the concern of the personal privacy. For example, the surveillance systems installed in the patient rooms monitor personal activities by analyzing their poses while at the same time, the personal appearance is inevitably disclosed to the systems. Therefore, predicting human poses using vision-based input encounters adverse lighting and privacy invasion issues.

\begin{figure}
    \centering
    \includegraphics[width=0.9\linewidth]{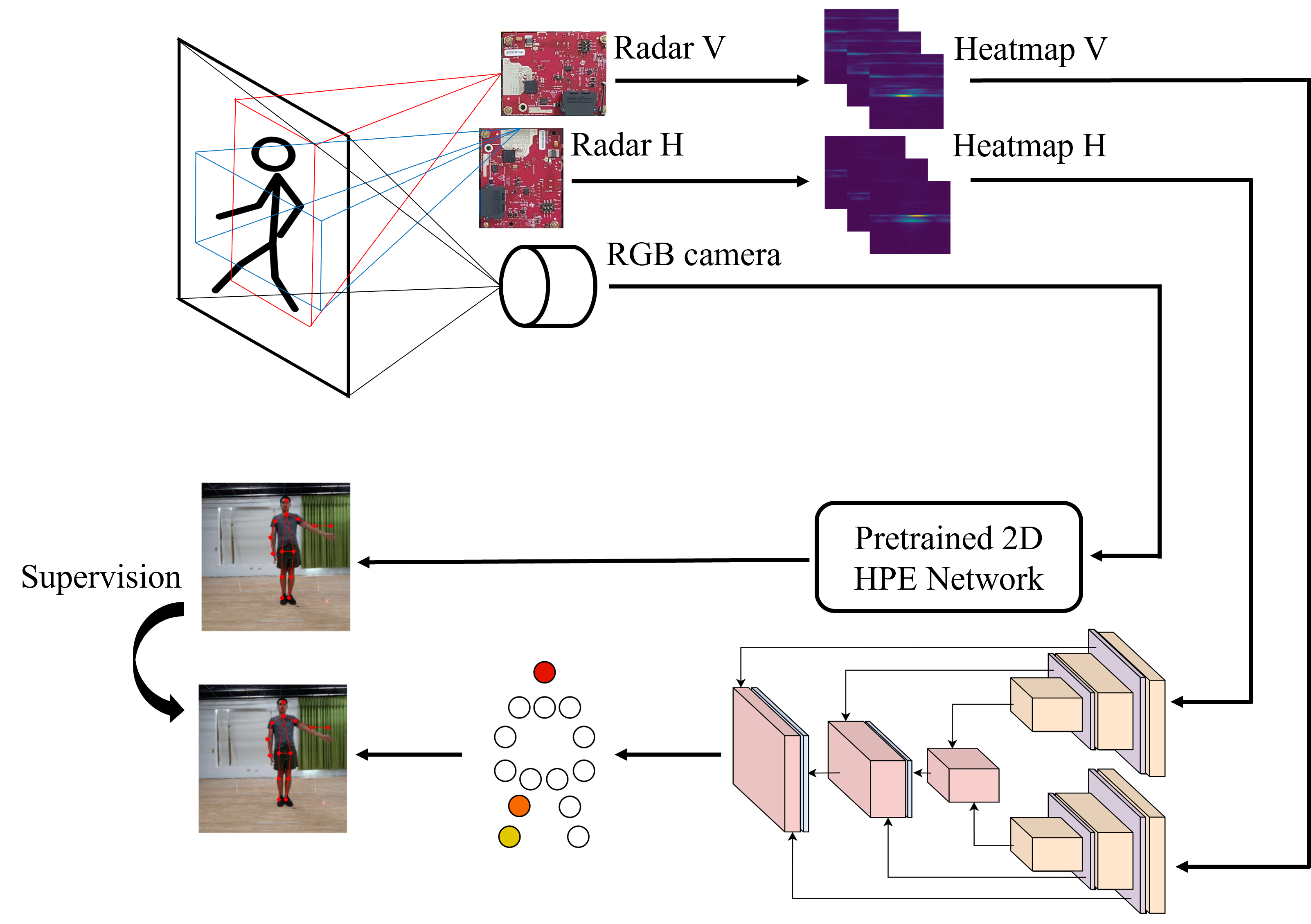}
    \caption{Illustration of our overall system setup. The upper branch represents mmWave Data Collection and Pre-processing. The lower branch represents our proposed Cross and Self Attention Module and  Pose Refinement Graph Convolutional Networks. H: Horizontal, V: Vertical.}
    \label{fig:Teaser}
\end{figure}
To address the above issues, a new type of HPE task has been proposed. 
Several radio frequency (RF) datasets ~\cite{rfpose, mmpose, captureHPE} are built to predict human skeletions.
Such RF signals are robust to lighting conditions and barely visually perceivable by humans. RF signals can be categorized by the frequency bands with varying characteristics. Zhao \etal~\cite{rfpose} adopt Wi-Fi signals (2.4 GHz), which possess a unique ability that it is able to capture a person's pose even when he or she is standing behind a wall. In spite of showing excellent results on 2D pose estimation, the Wi-Fi sensors used by~\cite{rfpose} is proprietorially-designed.
On the other hand, Sengupta \etal~\cite{mmpose} introduces another type of RF signal, Frequency Modulated Continuous Wave (FMCW) radar, with frequency band periodically changing from 77GHz to 81GHz, which can precisely detect the depth (range) and the velocity of an object. Comparing to the Wi-Fi sensor used by~\cite{rfpose}, the mmWave radar sensor is more economical and accessible, as well as commercially available from many instrument providers~\cite{TI}. The 3D HPE results shown in~\cite{mmpose} seem promising; however, it ignores the human body keypoints with high uncertainty, such as wrists, due to their low prediction accuracy, showing an inferior capability of capturing human poses using radar. Most importantly, the datasets of both \cite{rfpose, mmpose} remain inaccessible to the public, restricting the further development of an HPE in terms of RF data.

To overcome the challenging issues encountered in RGB-based and RF-based HPE, we introduce a new benchmark, Human Pose with mmWave Radar (HuPR). Unlike \cite{mmpose}, we additionally incorporate velocity information in our dataset, since radar sensors can provide a highly precise velocity information. Meanwhile, we propose a Cross- and Self-Attention Module (CSAM) to better fuse the multi-scale features from horizontal and vertical radars and a 2D pose confidence refinement network based on Graph Convolutional Network (PRGCN) to refine the confidence in the output pose heatmaps. Our framework consists of two cascading components, 1) Multi-Scale Spatio-Temporal Radar Feature Fusion with CSAM, which contains two branches to encode temporal range-azimuth and range-elevation information respectively, followed by a decoder to decode the fused features at every scale and predict 2D pose heatmaps and 2) PRGCN which is applied to the output heatmaps to refine the confidence of each keypoint based on a pre-defined graph of human skeletons. Our contributions are threefold:
\begin{itemize}
    \item We introduce a novel RF-based HPE benchmark, HuPR, which features privacy-preserving data, economical and accessible radar sensors, and handy hardware setup. The dataset and implementation code will be released upon paper acceptance.
    \item We propose a new radar pre-processing method that better extracts velocity information from radar signals to help RF-based HPE.   
    \item We propose CSAM to relate the features from two different radars for better feature fusion and PRGCN to refine the confidence of each keypoint, especially to improve the precision of the faster moving edge keypoints, such as wrists. Experimental results and ablation studies show that our proposed method makes significant improvement over RF-based 2D HPE methods and 3D pointcloud-based methods.
\end{itemize}


\section{Related Work}
\begin{figure}[t]
    \centering
    \setlength\tabcolsep{0.5em}
    \begin{tabular}{cccc}
        \includegraphics[width=0.2\linewidth]{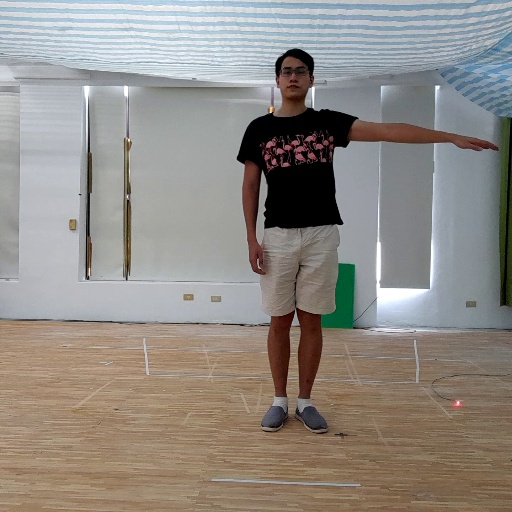} & \includegraphics[width=0.2\linewidth]{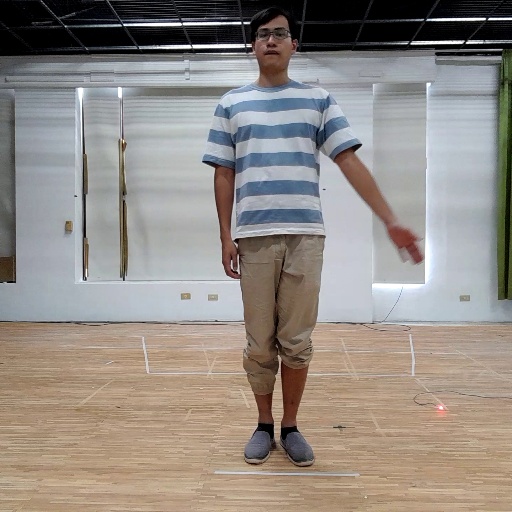} &
        \includegraphics[width=0.2\linewidth]{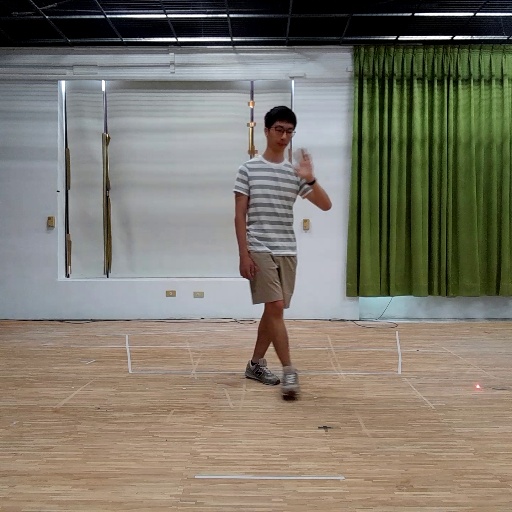} \\
        \includegraphics[width=0.2\linewidth]{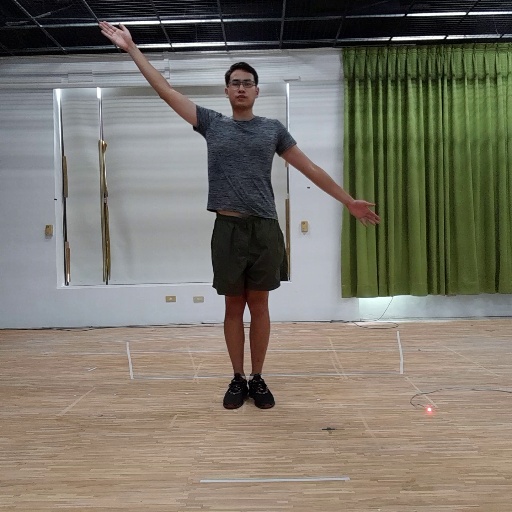} &
        \includegraphics[width=0.2\linewidth]{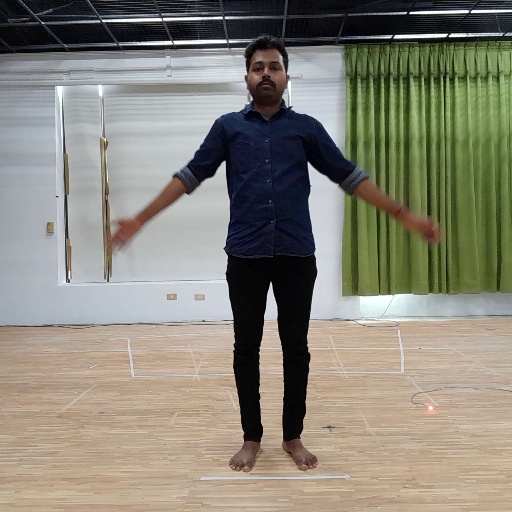} &
        \includegraphics[width=0.2\linewidth]{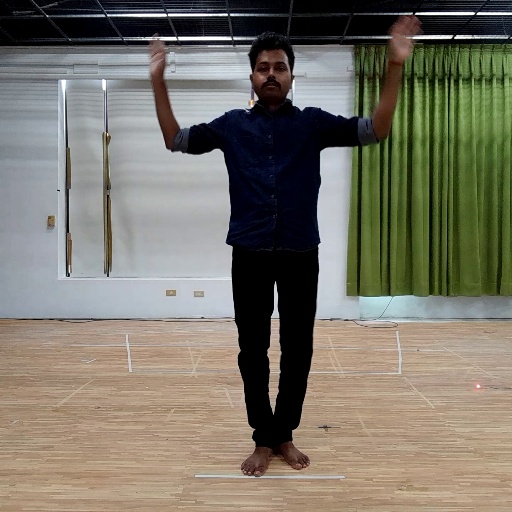} \\
        \includegraphics[width=0.2\linewidth]{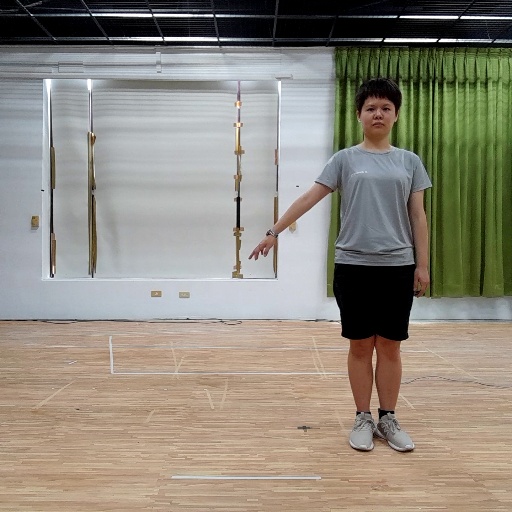} &
        \includegraphics[width=0.2\linewidth]{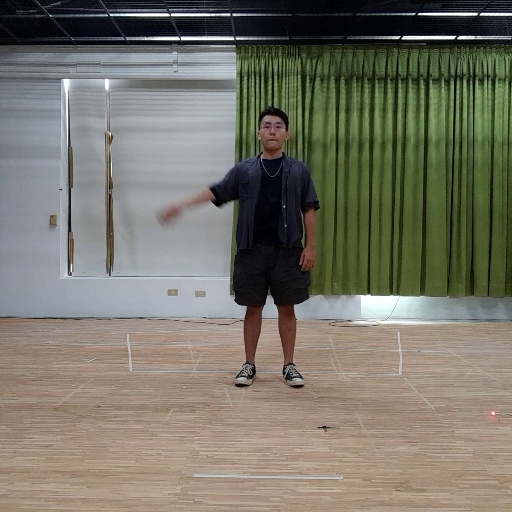} &
        \includegraphics[width=0.2\linewidth]{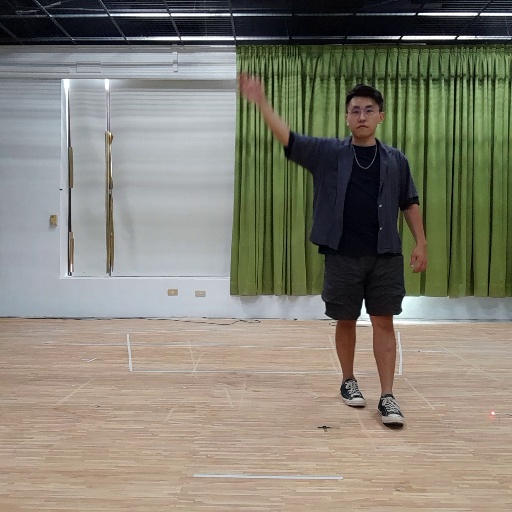} \\
    \end{tabular}
    
    \caption{Examples of actions in our dataset, including standing with fixed actions, standing with waving hands, and walking with waving hands.}
    \vspace{-0.8em}
    \label{fig:dataset_img}
\end{figure}
\subsection{RGB-based HPE}
There have been extensive  studies on RGB-based HPE. In general, these works can be split into two categories: regression-based methods and heatmap-based methods. Traditional regression-based methods \cite{SPM, PointSetNet, RLE} map input sources to the coordinates of body keypoints via an end-to-end neural network. The regression-based solutions are straightforward but less attractive since it is more difficult for a neural network to map image features into just several keypoint coordinates. On the other hand, heatmap-based methods~\cite{hrnet, SimplePose, hourglass, MaskRCNN, OpenPose} generally outperforming regression-based methods and dominate the field of HPE. Heatmap-based HPEs produce likelihood heatmaps for each keypoint as the target of pose estimation.

\subsection{RF-based HPE}
RF-based data are often used to deal with simpler human sensing tasks, such as activity recognition \cite{Radhar, DeepMV}, gesture recognition \cite{Learning_to_Sense, dfgr} and human object detection \cite{reid, PersonInWiFi}. Channel State Information (CSI) data are the main RF sources in early days, but they do not provide range or distance information. With the development of economical radio sensors, the estimation of range and angle of arrival becomes feasible with affordable devices, allowing more detailed and complicated tasks like HPE to be conducted on RF-based data~\cite{rfpose, mmpose, mmposeNLP, realtimeHuPose}. Zhao \etal~\cite{rfpose} utilize WiFi-ranged FMCW signals with the ability to generate the 2D human skeletons through walls, which exhibits the potential of privacy invasion. Sengupta \etal\cite{mmpose} use millimeter waves (mmWave) radars to collect the point clouds which are used for predicting 3D human skeletons. However, the edge keypoints like wrists are not considered. 

\subsection{Graph Convolutional Networks (GCN) in HPE}

Human skeletons inherently preserve graphical characteristics with its predefined relationship between keypoints. Many GCN-based solutions \cite{SpatialTemporalGCN, gcn3dpose, GraphPCNN, dgcn_aaai20} therefore well leverage this structural information and achieve a great success on HPE. Cai \etal~\cite{SpatialTemporalGCN} exploit spatio-temporal information to construct 3D poses by feeding 2D poses into a GCN. Wang \etal~\cite{GraphPCNN} refine several sampled human keypoint maps based on guided keypoints to locate the corresponding features. Qiu \etal~\cite{dgcn_aaai20}, similar to~\cite{GraphPCNN}, use predicted 2D keypoint coordinates to locate the features in the feature extractor for each keypoint, followed by a GCN to predict 3D keypoints. Our work directly utilizes the keypoint heatmaps as the node features, instead of the features associated with the keypoints, and then refines these heatmaps by a GCN.

\section{Dataset}

\begin{figure}
    \centering
    \includegraphics[width=0.5\linewidth]{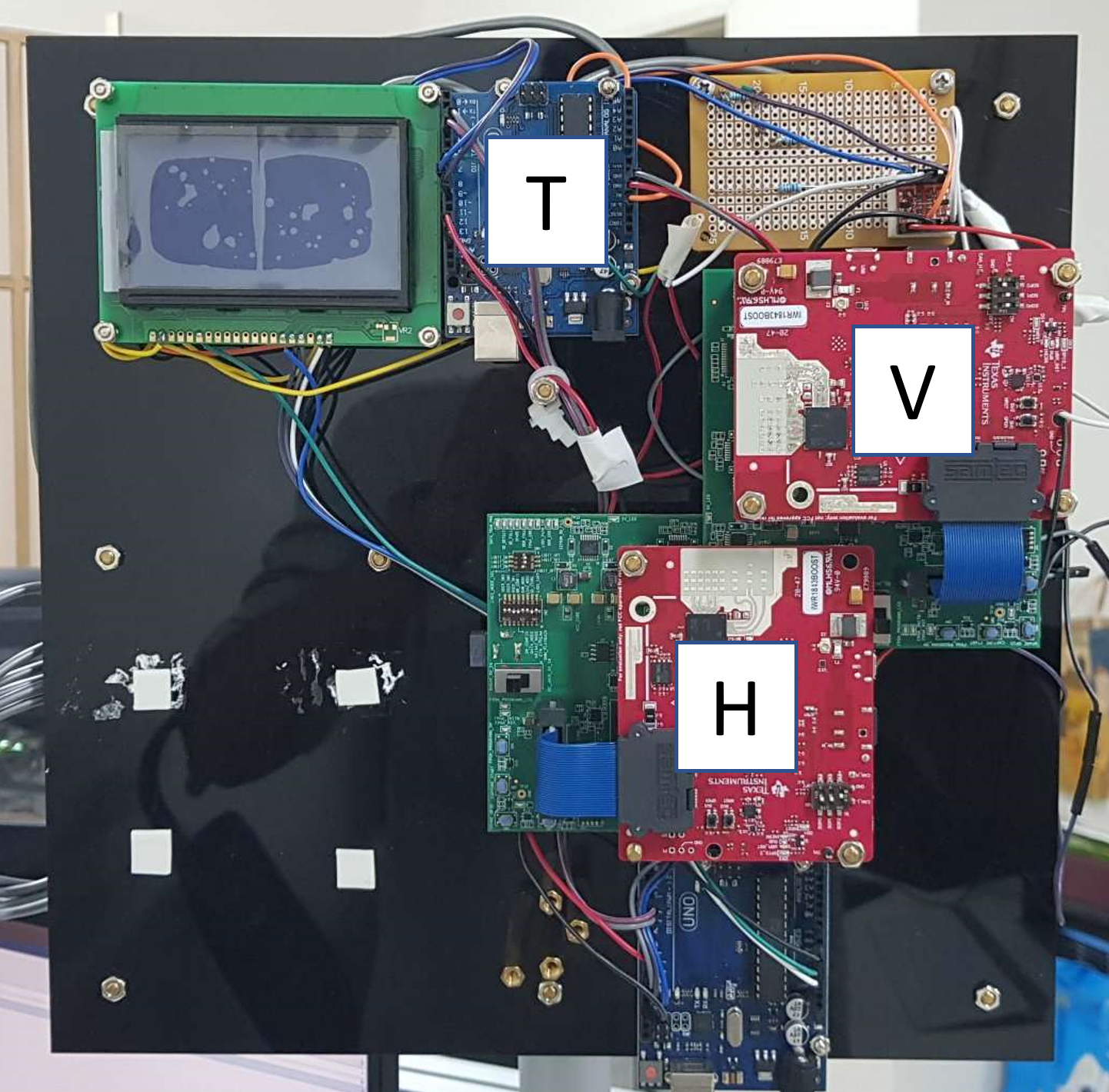}
    \caption{Our mmWave radar sensor board setup. H: Horizontal Radar Module, V: Vertical Radar Module, T: Trigger Generator}
    \vspace{-1.2em}
    \label{fig:dataset_radar}
\end{figure}

\subsection{Sensor Setup}

We use 2 identical radar sensors, IWR1843BOOST by Texas Instruments \cite{TI}. The radar sensors can provide the 3D information (range, azimuth, and elevation) of the scene; however, the resolution of elevation angle is extremely low (only 2 bins). Therefore, one radar sensor is rotated by 90\textdegree{} to the other radar sensor on the antenna plane. This way we have one radar sensor that focuses on the horizontal plane (range-azimuth-elevation) and another focuses on the vertical plane (range-elevation-azimuth). The physical setup of our radar sensors is shown in Fig.~\ref{fig:dataset_radar}, and an illustration of data acquiring process is shown in Fig.~\ref{fig:Teaser}. Both 
radar sensors are fixed on a solid plane and kept as near as 3cm from each other. One RGB camera is fixed at the same plane. RGB frames from the camera are used to generate the 2D ground-truth keypoints based on a pre-trained image-based 2D human pose estimation network  HRNet~\cite{hrnet}.

for the IWR1843BOOST module the maximum range is set to 11m with the range resolution of 4.8cm, the azimuth angle of 120\textdegree{} with the resolution of 30\textdegree{}, and an elevation angle of 30\textdegree{} with the resolution of 15\textdegree{}. To increase the azimuth resolution, we configure the radar sensor in 1 virtual transmitter antenna and 8 virtual receiver antennas, which in turn increase the azimuth resolution by a factor of 2, resulting in the final azimuth resolution of 15\textdegree{}. Table~\ref{table:radar_settting} shows the parameter settings of our radar.

\begin{table}[t]
\centering
\setlength\tabcolsep{2em}
\caption{Settings and properties of our radar sensors}
\vspace{-0.6em}
 \begin{tabular}{lc} 
 \toprule
 Parameters & Values \\
 \midrule
 Physical Transmitters & 4 \\ 
 Physical Receivers & 3 \\
 Frequency Slope & 60.012 MHz \\
 Sample Rate & 4400 ksps \\
 Chirp Duration & 72 us \\
 Chirps in Frame & 64 \\
 ADC Samples & 256 \\
 Frame Per Second & 10 \\
 \bottomrule
 \end{tabular}
 \label{table:radar_settting}
\end{table}

\begin{figure*}[t]
    \centering
    \includegraphics[width=\linewidth]{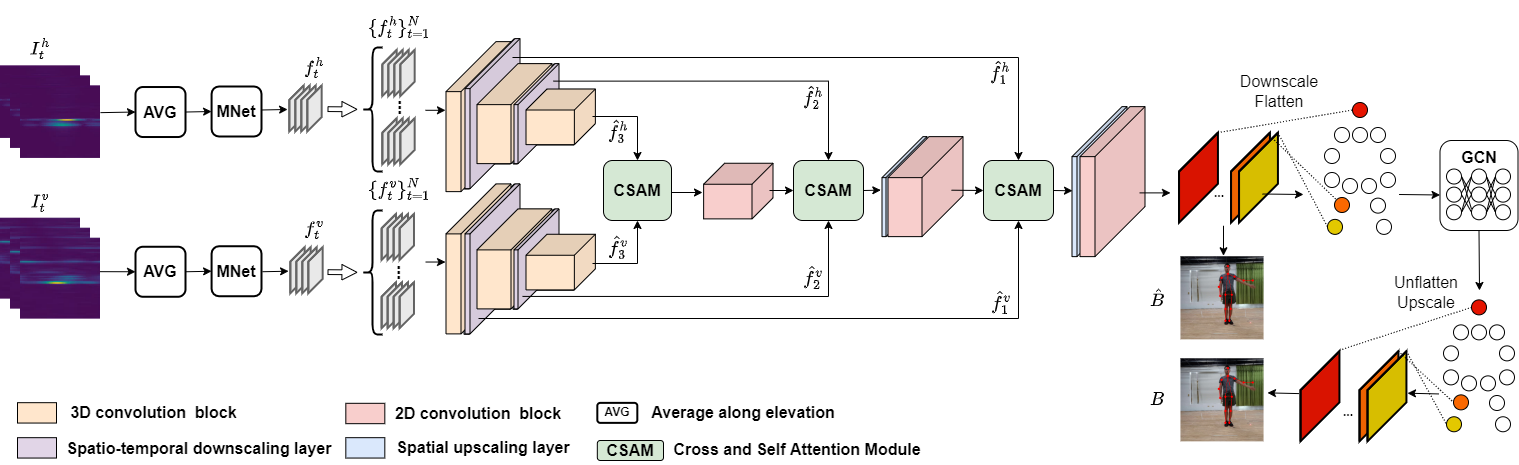}
    \caption{Architecture of our proposed method.}
    \label{fig:architecture}
\end{figure*}

\subsection{Dataset Collection}

We collected 235 sequences of data in an indoor environment, with each sequence being one-minute long and totally about 4 hour-long video data. Some samples in the dataset are shown in Fig~\ref{fig:dataset_img}. Each sequence has an RGB camera frames, the horizontal radar frames, and the vertical radar frames. The two radars and the camera are synchronously configured to capture 10 frames per second (FPS), and hence each sequence has 600 triplets of camera-radar-radar frames. In total, we have 141,000 triplets in 235 sequences. The resolution of camera frames is $256 \times 256$ and the dimensions of raw radar data is $256 \times 64 \times 8 \times 2$, corresponding to analog-to-digital converter (ADC) samples, chirps, azimuth bins, and elevation bins, respectively. Specifically, we select a specific range of the ADC samples to just process signals in a shorter range since human subjects will always perform actions close but not too closed to the sensors and use zero paddings to increase the size of the azimuth and elevation dimension before the pre-processing procedure. Overall, we obtain the raw radar data with the dimensions of $64 \times 64 \times 64 \times 8$.
There is always only one person in the scene. The person performs 3 types of movements: static actions, standing and waving hand(s), and walking with waving hand(s).
Static actions mean the subject performs a pose without moving for at least 10 seconds.
Six subjects are involved in the dataset and the consent of each subject has been obtained by a face-to-face interview. One of them only appears in the test set for fair evaluation.

\begin{figure}
\centering
\begin{subfigure}{0.4\textwidth}
    \includegraphics[width=\textwidth]{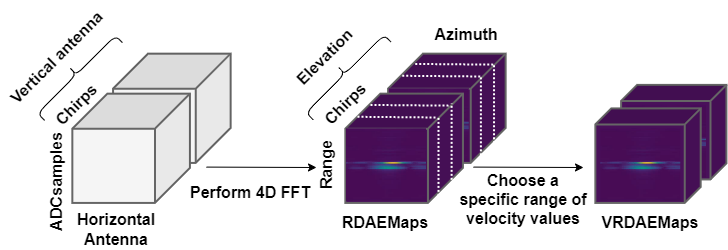}
    \caption{}
    \label{fig:first}
\end{subfigure}
\begin{subfigure}{0.4\textwidth}
    \includegraphics[width=\textwidth]{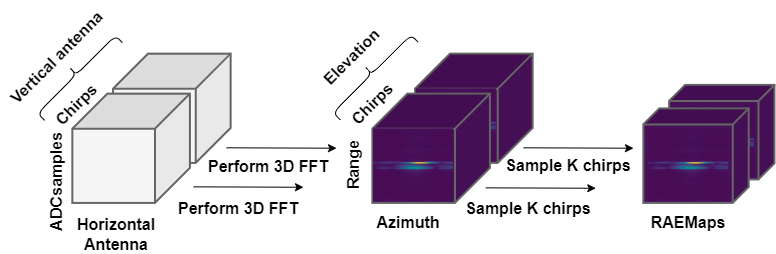}
    \caption{}
    \label{fig:second}
\end{subfigure}
\hfill

\caption{Comparison of radar pre-processing methods: (a) our proposed method and (b) traditional schemes.}
\label{fig:preprocessing}
\end{figure}

\section{Proposed Method}\label{sec:method}
This work addresses the task of 2D human pose estimation using mmWave radars. Given two input sequences of length $N$ consisting of the two sets of radar frames $\{I^h_t\}^N_{t=1}$ and $\{I^v_t\}^N_{t=1}$ (to be defined in Section \ref{subsec:rsp}) from the horizontal and the vertical radar sensors, where $I^h_t, I^v_t \in \mathbb{R}^{2 \times K \times H \times W \times E}$, our task is to predict the 2D keypoint heatmaps $B \in \mathbb{R}^{C \times H \times W}$ for the scene at time instance $t = \frac{N}{2}$. Each keypoint's heatmap $B_c,c \in C$ represents the confidence distribution of the keypoint $c$. In particular, we take a sliding window approach to include radar frames from not only the past but also the future when the prediction proceeds from one time instance to the next. 
Fig.~\ref{fig:architecture} shows the overall pipeline of our proposed framework. The two input sequences $\{I^h_t\}^N_{t=1}$ and $\{I^v_t\}^N_{t=1}$ are first processed by a Multi-Scale Spatio-Temporal Feature Fusion module to generate the 2D keypoint heatmaps $\hat{B} \in \mathbb{R}^{C \times H \times W}$. The heatmaps $\hat{B}$ are then refined by Pose Refinement Graph Convolutional Networks (Section~\ref{subsec:prgcn}) for pose confidence refinement to produce the final heatmap $B$.

\subsection{Radar Signal Pre-processing}\label{subsec:rsp}
Fig.~\ref{fig:first} depicts our proposed radar pre-processing method for each radar frame, $I^h_t$ or $I^v_t$. Our method is designed to additionally extract velocity information as an important cue for predicting human pose.
With the traditional scheme (Fig.~\ref{fig:second}), the range-azimuth-elevation maps (RAEMap) are often generated by performing Fast Fourier transform (FFT) first along the dimension of digitized chirp samples (i.e. ADCsamples), followed by applying the other two FFT's to co-located samples along the horizontal and vertical antenna dimensions of the 3D raw input data tensor to get range, azimuth, and elevation information. The final RAEMaps are formulated by sampling uniformly $K$ chirps along the chirp dimension. Some of the previous methods~\cite{RODNet, captureHPE} adopt only range-azitmuth maps (RAMap) without elevation information. 
In comparison to the traditional scheme, our proposed pre-processing method generates the velocity-specific range-doppler-azimuth-elevation map (VRDAEMap) representation by additionally performing FFT along the chirp dimension, to extract the doppler velocity information. Specifically, we first perform 4D FFT on the raw data along all four dimensions of ADCsamples, chirps, horizontal antenna, and vertical antenna to obtain the range-doppler-azimuth-elevation map (RDAEMap). Instead of directly using the RDAEMap, which is sparse and inefficient to process, we choose a specific range of the velocity values (named VRDAEMap) from $\frac{-K}{2}$ to $\frac{K}{2}$ relative to the radar sensors as our input $I^h_t,I^v_t$ with the dimensions of $2 \times K \times H \times W \times E$, where $2$ represents the real and imaginary parts of the FFT coefficients. The values of $H$, $W$, and $E$ are $64, 64, 8$, respectively. We empirically set $K$ to 8 to account for the moderately slow action sequences in our applications.
\begin{figure}
    \centering
    \includegraphics[width=0.9\linewidth]{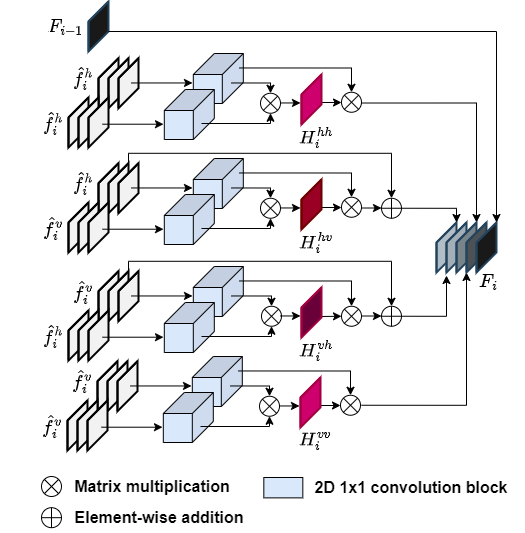}
    \caption{Architecture of Cross and Self Attention Module.}
    \label{fig:CSAM}
\end{figure}
\subsection{Multi-Scale Spatio-Temporal Radar Feature Fusion with Cross- and Self-Attention}
Our multi-scale spatio-temporal feature fusion is to fuse information captured by the horizontal and vertical radars, in generating the heatmaps of 2D keypoints. To this end, we first process two sequences of the VRDAEMaps $\{I^h_t\}^N_{t=1}$ and $\{I^v_t\}^N_{t=1}$ by two distinct branches yet with the same network structure. As shown in the left part of Fig.~\ref{fig:architecture}, each branch begins with an averaging (AVG) function that pools information along the elevation dimension (dimension $E$) by taking the average of co-located feature samples. The intermediate result of dimension $2 \times K \times H \times W$ is processed further by a fusion network, MNet~\cite{RODNet}, which fuses the information along the velocity dimension (dimension $K$) using a convolutional layer with a kernel size $\frac{K}{2} \times 1 \times 1$ and a MaxPooling layer of stride $\frac{K}{2}$ along the velocity dimension. At the output of MNet~\cite{RODNet}, the number of feature channels is set to $D$. We therefore get two sets of fused feature maps $\{f^h_t\}^N_{t=1}$ and $\{f^v_t\}^N_{t=1}$, where $f^v_t, f^h_t \in \mathbb{R}^{D \times H \times W}$.

\textbf{Multi-Scale Spatio-temporal Feature Fusion:} The middle part of Fig.~\ref{fig:architecture} illustrates our multi-scale spatio-temporal feature fusion. We adopt multiple 3D convolutional layers with strides to aggregate information with respect to both the spatial and the temporal dimensions. 
The output of each spatial scale is followed by a 3D convolutional layer with a kernel size $\frac{N}{2^{i-1}} \times 1 \times 1$ to aggregate the remaining temporal information, resulting in the encoding features $\hat{f}^v_i, \hat{f}^h_i$, where $i$ is the scale index and $\hat{f_i}^v, \hat{f_i}^h \in \mathbb{R}^{2^{i}\cdot D \times \frac{H}{2^{i-1}} \times \frac{W}{2^{i-1}}}$. The encoding features of each scale are processed by a Cross- and Self-Attention module to generate the decoding features. We then use multiple 2D convolutional layers, followed by a sigmoid function, to generate 2D keypoint heatmaps $\hat{B}$.

\textbf{Cross- and Self-Attention Module:}
Our cross- and self-attention module (CSAM) is to discover useful context information for human pose detection across and within the radar frames captured by the horizontal and vertical radars. It performs cross-attention and self-attention at different scales of the horizontal and vertical radar features. 
We first describe the cross-attention mechanism (the central part of Fig.~\ref{fig:CSAM}). It begins by applying 2D 1x1 convolution to $\hat{f}^h_i$ and $\hat{f}^v_i$ independently to generate the projected features. The resulting features are then flattened along the spatial dimensions to obtain the attention maps $H^{hv}_i, H^{vh}_i$, where $H^{hv}_i\in \mathbb{R}^{\frac{HW}{2^{i-1}\times2^{i-1}} \times \frac{HW}{2^{i-1}\times2^{i-1}}}$ (respectively, $H^{vh}_i$) indicates the use of $\hat{f}^v_i$ (respectively, $\hat{f}^h_i$) as the guiding signal to formulate the attention map with $\hat{f}^h_i$ (respectively, $\hat{f}^v_i$) serving as the reference signal. Finally, $\hat{f}^h_i$ is updated (respectively, $\hat{f}^v_i$) by performing cross-attention with $H^{hv}_i$ (respectively, $H^{vh}_i$) to generate the cross-attended residual features. Note that the width (dimension $W$) of horizontal and vertical features represent different meanings (azimuth and elevation information, respectively). Directly relating two features may make the training process unstable. We therefore add a residual connection at the end of the layer to stabilize the whole training process.
The self-attention mechanism (the top and bottom branches of Fig.~\ref{fig:CSAM}) acts in a way similar to the cross-attention one. Two major distinctions are that both the guiding and reference signals come from the same $\hat{f}^h_i$ or $\hat{f}^v_i$, and that there is no skip connection to generate the self-attended residual features. Finally, we concatenate all the output features, generating $F_i$ as the input to the convolution blocks and CSAM of the next scale.
\begin{table}[t]
    \centering
    \setlength\tabcolsep{0.4em}
    \caption{Comparison of pre-processing methods in $AP$. Total denotes the average precision over 14 keypoints.}
    \vspace{-0.6em}
    \begin{tabular}{ll|ccc}
    \toprule
    Pre-processing & Model & Elbow & Wrist  & Total \\
    \midrule
    RAEMap & RF-Pose~\cite{rfpose} & 13.6 & 6.0 & 40.6\\
    RAEMap & Ours & 40.9 & 17.4 & 61.6 \\
    VRDAEMap & Ours & \textbf{45.6} & \textbf{23.5} & \textbf{64.3} \\
    \bottomrule
    \end{tabular}
    \label{table:preproc_results}
    \vspace{-0.3cm}
\end{table}

\begin{table*}[t]
    \centering
    \setlength\tabcolsep{0.4em}
    \caption{Comparison of baseline and the variants of our proposed method. CSAM denotes that we train the model with cross-attention (CA) + and self-attention (SA) module. $\dagger$ indicates the results are reproduced by us.}
    \vspace{-0.6em}
    \begin{tabular}{l|cccccccc|ccc}
    \midrule
     & \multicolumn{8}{c}{$AP$} & $AP$ & $AP^{50}$ & $AP^{75}$ \\
    Model & Head & Neck & Shoulder & Elbow & Wrist & Hip & Knee & Ankle & \multicolumn{3}{c}{Total}\\
    \midrule
    RF-Pose$\dagger$~\cite{rfpose} & 61.0 & 65.3 & 52.5 & 16.1 & 6.3 & 73.5 & 65.7 & 62.0 & 41.4 & 82.9 & 37.0 \\ 
    
    Ours & 71.3 & 76.7 & 64.4 & 36.9 & 18.8 & 84.8 & 78.1 & 70.4 & 57.6 & 96.5 & 64.1 \\
     
    
    Ours (CA) & \textbf{79.6} & \textbf{83.8} & \textbf{73.3} & 39.3 & 16.7 & 84.3 & 80.6 & 73.1 & 60.0 & 96.8 & 69.7 \\ 
     
    
    Ours (SA) & 76.4 & 81.6 & 71.4 & 42.1 & 18.7 & 87.2 & 81.7 & 72.4 & 61.3 & 96.9 & 71.2 \\ 
     
    
    Ours (CSAM) & 74.4 & 80.8 & 69.0 & 42.1 & 17.4 & 87.2 & \textbf{82.4} & \textbf{75.6} & 61.9 & \textbf{97.3} & 73.1 \\ 
     
     
    Ours (CSAM + PRGCN) & 77.5 & 81.9 & 70.3 & \textbf{45.5} & \textbf{22.3} & \textbf{88.1} & 82.2 & 73.1 & \textbf{63.4} & 97.0 & \textbf{74.0} \\
    \bottomrule
    \end{tabular}
    \label{table:comparison_results}
\end{table*}

\begin{table}[t]
    \centering
    \setlength\tabcolsep{0.4em}
    \caption{Comparison of 3D keypoint performance based on MPJPE in millimeters. Ours + VideoPose3D means that we adopt our proposed method to generate 2D keypoints, which are lifted to 3D by VideoPose3D.}
    \vspace{-0.6em}
    \begin{tabular}{l|ccc}
    \toprule
    Model & Elbow & Wrist & Total \\
    \midrule
    mmMesh~\cite{mmMesh} & 112.9 & 218.2 & 71.3\\
    Ours + VideoPose3D~\cite{videopose3d} & \textbf{85.3} & \textbf{156.4} & \textbf{68.2} \\
    \bottomrule
    \end{tabular}
    \label{table:3d_results}
\end{table}

\subsection{Pose Refinement Graph Convolutional Networks (PRGCN)}\label{subsec:prgcn}
The locations of different body keypoints are strongly related to each other. That is, knowing the locations of some keypoints is able to help predict the locations of the others. This motivates us to develop a pose refinement module based on GCN to refine the initial keypoint heatmaps $\hat{B}$. We first define the graph structure. Let $\mathcal{G} = (\mathcal{V}, \mathcal{E})$ be a graph, where $\mathcal{V} = \{v_1, ..., v_C\}$ is a collection of nodes, each corresponding to a keypoint to be estimated. $\mathcal{E} = \{A_{i,j} | v_i, v_j \in \mathcal{V}\}$ specifies edges connecting nodes $v_i$ and $v_j$ with the weight $A_{i,j} = 1$ if the two body keypoints $i,j$ are physically connected or $0$ otherwise. In our task, $v_i \in \mathbb{R}^{\frac{H \times W}{4}}$ represents the downscaled heatmap of the $i$-th keypoint. 
It is obtained by downscaling and flattening the final features $\Bar{f} \in \mathbb{R}^{C \times H \times W}$ (see Fig.~\ref{fig:architecture}) along the spatial dimensions into a total of $C$ 1-dimensional vectors, with each being a node feature $v_i,i \in C$.
To generate the refined keypoint heatmaps $B$, we perform the feature propagation and the inference on graph $\mathcal{G}$ through a 3-layer GCN. The forward pass proceeds as follows:
\begin{equation}\label{eq:gcn}
    \begin{split}
        V' = \sigma_r{(\hat{A}VW_1)}, \\
        V'' = \sigma_r{(\hat{A}V'W_2)}, \\
        H^g = \sigma_s{(\hat{A}V''W_3)}
    \end{split}
\end{equation}
where $V = [v_1, ..., v_C]^T \in \mathbb{R}^{C \times \frac{H \times W}{4}}$ is a matrix of node features; $W_1, W_2, W_3 \in \mathbb{R}^{\frac{H \times W}{4} \times \frac{H \times W}{4}}$ are learnable parameters; $\sigma_r{(\cdot)}, \sigma_s{(\cdot)}$ are ReLU and sigmoid functions, respectively; and $\hat{A} = A + I$ is the sum of the adjacency matrix $A \in \mathbb{R}^{C \times C}$ and an identity matrix $I$. The refined output $B$ is un-flattened back to heatmaps with dimensions $\frac{H}{2} \times \frac{W}{2}$, followed by upscaling. 
Note that the previous GCN-based approach \cite{gcn3dpose} produces the node features by selecting the co-located feature samples of intermediate features at the coordinates of keypoints. Our work, however, does not follow the same feature selection process. The reason is that the radar signals and the keypoint heatmaps have different coordinate systems. We are unable to map radar features using keypoint coordinates. To train the model end-to-end, we impose pixel-wise binary cross-entropy on both the initial keypoint heatmaps $\hat{B}$ and the GCN-refined keypoint heatmaps $B$. Regularizing $\hat{B}$ makes the training of $B$ more stable since its inputs depend completely on $\hat{B}$. To be specific, the objective function is given by
\begin{equation}\label{eq:loss}
    \mathcal{L} = \mathcal{L}_{bce}(\hat{B}, T) + \alpha \cdot \mathcal{L}_{bce}(B, T)
\end{equation}
where $\alpha$ is a hyper-parameter and is set to $1$ in our implementation. Taking $\mathcal{L}_{bce}(B,T)$ as an example, the binary cross entropy loss is defined as: 
\begin{equation}\label{eq:loss}
\begin{aligned}
&\mathcal{L}_{bce}(B, T) = \\ 
&- \sum_{c, i, j}T_{c, i, j}\log{(B_{c, i, j})} + (1-T_{c, i, j})\log{(1 - B_{c, i, j})}
\end{aligned}
\end{equation}
where $T \in \mathbb{R}^{C \times H \times W}$ is the heatmaps of ground-truth  keypoints generated with HRNet \cite{hrnet}. Note that the heatmap of each keypoint is a Gaussian distribution with mean at the ground-truth keypoint coordinates and a pre-determined variance.
\section{Experimental Results}

\subsection{Setup and Evaluation Metrics}
We train the proposed system on our HuPR dataset, which contains total 235 video sequences, each being 1-minute long. We take 193 video sequences for training, 21 for validation, and 21 for test. For fair comparison, we only choose RF-Pose~\cite{rfpose} with RF-based input data as our heatmap-based 2D HPE baseline since there are only a handful of prior works, most of which do not even release nor provide hardware settings and training data. It is challenging to reproduce their results. We additionally include mmMesh~\cite{mmMesh}, which is a pointcloud-based 3D mesh prediction network. We modified it to 3D HPE network, to see how our RF-based scheme performs as compared to the pointcloud-based scheme. Following common practice~\cite{rfpose, mscoco}, we adopt average precision over different object keypoint similarity (OKS) as the performance metric of 2D keypoints. Specifically, we evaluate the model over 14 keypoints, including head, neck, shoulders, elbows, wrists, hips, knees, and ankles. We present three variants of average precision $AP^{50}$, $AP^{75}$, and $AP$. The value of 50 and 75 indicate the loose and strict constraints of OKS. $AP$ denotes the average precision over 10 OKS thresholds, namely, 0.5, 0.55, ..., 0.9, 0.95. Following~\cite{videopose3d}, we report mean per-joint position error (MPJPE) in millimeters as the evaluation metric of 3D keypoints.

\subsection{Implementation Details}
We implement our pre-processing method and the proposed framework with Python and Pytorch. The 2D ground-truth keypoints for training video sequences are generated from the RGB frames by the state-of-the-art 2D human pose estimation network, HRNet~\cite{hrnet}, which is pre-trained on MPII dataset \cite{mpii}. 
We also choose the keypoints generated by HRNet as the 2D pseudo ground-truths for our test set. We remark that our dataset contains mostly simple actions under a single scene. A side experiment shows that HRNet achieves an $AP$ of 99\%. In other words, HRNet provides satisfactorily good ground-truths under our application scenarios. 
Likewise, to generate the 3D ground-truths of the training and test sets, we choose VideoPose3D~\cite{videopose3d}. Since our method generates only 2D keypoints, we adopt the pre-trained network of VideoPose3D~\cite{videopose3d} to convert them into 3D keypoints.
We use the Adam optimizer with a learning rate 0.0001, which is decreased by a factor of 0.999 for every 2000 iterations. We set the batch size to 20 and the weight decay to 0.0001. The number ($N$) of input frames is chosen to be 8 frames (0.8 seconds) for our method, and 24 frames (2.4 seconds) for RF-Pose~\cite{rfpose} which shows the best results on our dataset. 
Because the code of RF-pose~\cite{rfpose} is unavailable, we implement it by following the network structure reported in the paper. In addition, we train the model from mmMesh~\cite{mmMesh} from scratch using their official code. In particular, we replace the mesh reconstruction module with MLPs to serve our purpose of predicting 3D keypoints.
Each convolutional block in Fig.~\ref{fig:architecture} contains two basic blocks from ResNet~\cite{resnet}. The activation functions used in the 3D and 2D blocks are ReLU and PReLU, respectively. Batch normalization is not included in the 2D blocks. In these basic blocks, the 2D convolutional layer has a kernel size $3 \times 3$, and the 3D one has a kernel size $3 \times 3 \times 3$. We use a single Tesla V100 to train the networks.
\subsection{Quantitative Results}
\textbf{Comparison of radar pre-processing methods:}
Table~\ref{table:preproc_results} shows the results for two pre-processing methods. These methods are trained on a simplified dataset, containing 80 video sequences for training, 10 for validation, and 10 for test. Comparing with the RAEMap, the traditional method, our proposed VRDAEMap achieves higher $AP$, especially on difficult keypoints like wrists and elbows.
We find that RF-Pose~\cite{rfpose} shows relatively poor overall accuracy when compared with RAEMap (Ours). As RF-Pose~\cite{rfpose} does not consider the chirp or velocity information in the Wi-Fi data, which plays an important role in the radar data, it is unable to take full advantage of the temporal information despite using a longer sequence of frames. 

\textbf{Comparison of methods:}
Table~\ref{table:comparison_results} compares the values of average precision of different methods. Our method (CSAM + PRGCN) outperforms RF-Pose~\cite{rfpose} in all $AP$ values. It also attains 5.8\% and 9.9\% gain in terms of $AP$ and $AP^{75}$, as compared to our baseline method (Ours). The model "Ours" denotes that we only perform concatenation of features of different scales from two radars without CSAM and PRGCN for refinement. The increase in $AP^{75}$, which is a stricter metric, indicates that our proposed method (CSAM + PRGCN) is able to predict the keypoints more precisely. We further analyze the results of the variants of our method in Section~\ref{subsection:abl}. Table~\ref{table:3d_results} further compares the 3D keypoint detection of our method to mmMesh~\cite{mmMesh}, i.e. the pointcloud-based state-of-the-art method. It is seen that our method outperforms mmMesh~\cite{mmMesh}, achieving 27.6, 61.8, and 3.1 less MPJPE of elbows, wrists, and total, respectively. Note that the results for mmMesh~\cite{mmMesh} are produced by training their model end-to-end to predict 3D keypoints. In comparison, our scheme predicts 3D keypoints in 2 sequential steps (i.e. in a non-end-to-end optimized manner): the generation of 2D keypoints followed by converting them into 3D keypoints using VideoPose3d~\cite{videopose3d}. We clarify that as the network of VideoPose3D~\cite{videopose3d} is pre-trained on another dataset,  it possesses the ability to improve the performance of 3D HPE by lifting the 2D predictions. In other words, it contributes partly to our superior performance which makes the comparison unfair to some extent.    

\begin{table}[t]
    \centering
    \caption{Comparison of different range of velocity values.}
    \vspace{-0.6em}
    \setlength\tabcolsep{1em}
    \begin{tabular}{c|ccc}
        \toprule
        & \multicolumn{3}{c}{$AP$} \\
        K & Elbow & Wrist & Total \\
        \midrule
        2 & 33.9 & 13.7 & 53.1 \\
        4 & 36.7 & 14.6 & 58.9 \\
        8 & \textbf{45.5} & \textbf{22.3} & \textbf{63.4} \\
        16 & 39.2 & 20.4 & 60.6 \\
        \bottomrule
    \end{tabular}
    \vspace{-0.6em}
    \label{table:ablation}
\end{table}


\subsection{Ablation Experiments}\label{subsection:abl}
\textbf{Component breakdown:}
We investigate the effect of each component in our proposed method. 
It is seen in Table~\ref{table:comparison_results} that with cross-attention only (CA), self-attention only (SA), or cross- and self-attention (CSAM), the predictions related to the torso, e.g. head, neck, and shoulders, become more stable.  Moreover, adopting PRGCN further improves the detection performance of wrists and elbows (i.e. fast-moving keypoints) up to 4.9\% and 3.4\% respectively. The reason may be attributed to that the pre-defined adjacency matrix propagates only the neighboring features to refine the final confidence output.   

\textbf{Velocity range:} 
We study how the velocity information derived from radar signals affect the performance of our pre-processing scheme. Table \ref{table:ablation} shows that increasing the velocity range, as specified by $K$ in Section~\ref{subsec:rsp}, is able to capture the movement of the edge keypoints more precisely. However, the larger velocity range makes the VRDAMaps become larger and sparser, leading to increased network complexity and memory footprint. In particular, when $K$ goes beyond 8, the performance declines. This may be because an extremely large $K$ value involves more noise rather than useful velocity information. 


\begin{figure}[t]
    \centering
    \setlength\tabcolsep{0.1em}
    \begin{tabular}{cccccccc}
         \raisebox{2.0\normalbaselineskip}[0pt][0pt]{\rotatebox[origin=c]{90}{GT}} & 
         \includegraphics[width=0.22\linewidth]{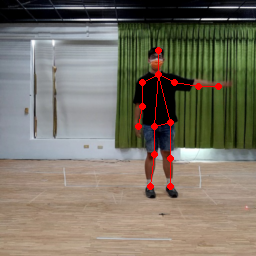}& 
         \includegraphics[width=0.22\linewidth]{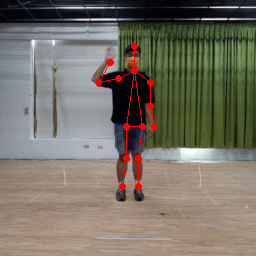}&
         \includegraphics[width=0.22\linewidth]{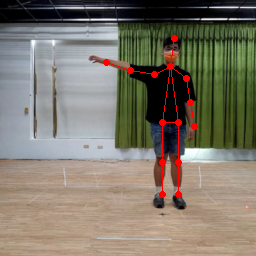}&
         \includegraphics[width=0.22\linewidth]{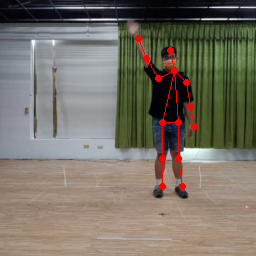}\\
         
         \raisebox{2.0\normalbaselineskip}[0pt][0pt]{\rotatebox[origin=c]{90}{RF-Pose~\cite{rfpose}}} & 
         \includegraphics[width=0.22\linewidth]{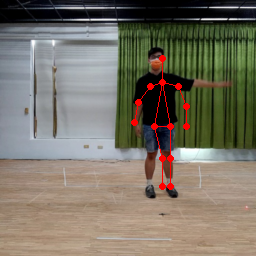} &
         \includegraphics[width=0.22\linewidth]{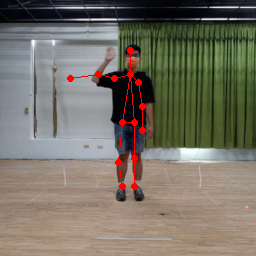} &
         \includegraphics[width=0.22\linewidth]{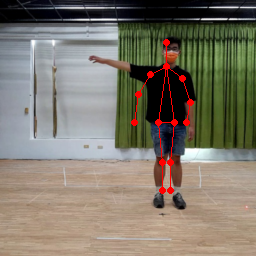} &
         \includegraphics[width=0.22\linewidth]{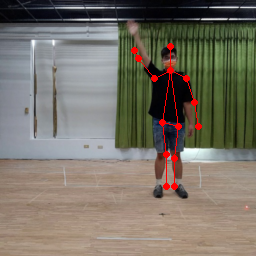} \\
         
         \raisebox{2.0\normalbaselineskip}[0pt][0pt]{\rotatebox[origin=c]{90}{Ours}} & 
         \includegraphics[width=0.22\linewidth]{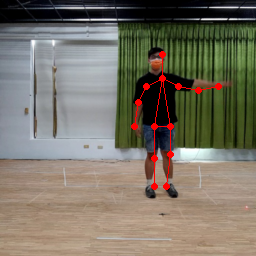} &
         \includegraphics[width=0.22\linewidth]{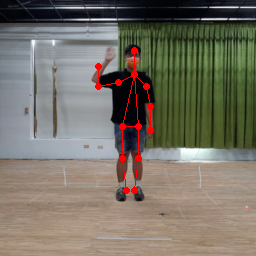} &
         \includegraphics[width=0.22\linewidth]{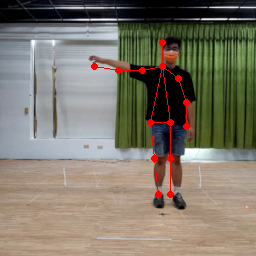} &
         \includegraphics[width=0.22\linewidth]{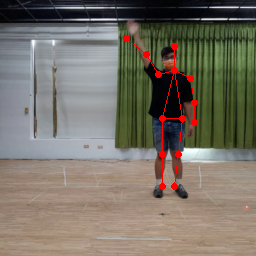} \\
    \end{tabular}
    \vspace{-0.8em}
    \caption{Comparison of 2D qualitative results. GT: Ground-truth keypoints generated by HRNet~\cite{hrnet}. Ours: The predicted results generated by our method (CSAM + PRGCN).}
    \label{fig:comp_qualitative}
\end{figure}
\begin{figure}[t]
    \centering
    \setlength\tabcolsep{0.1em}
    \begin{tabular}{cccccccc}
         \raisebox{2.0\normalbaselineskip}[0pt][0pt]{\rotatebox[origin=c]{90}{GT}} & 

         \includegraphics[width=0.32\linewidth]{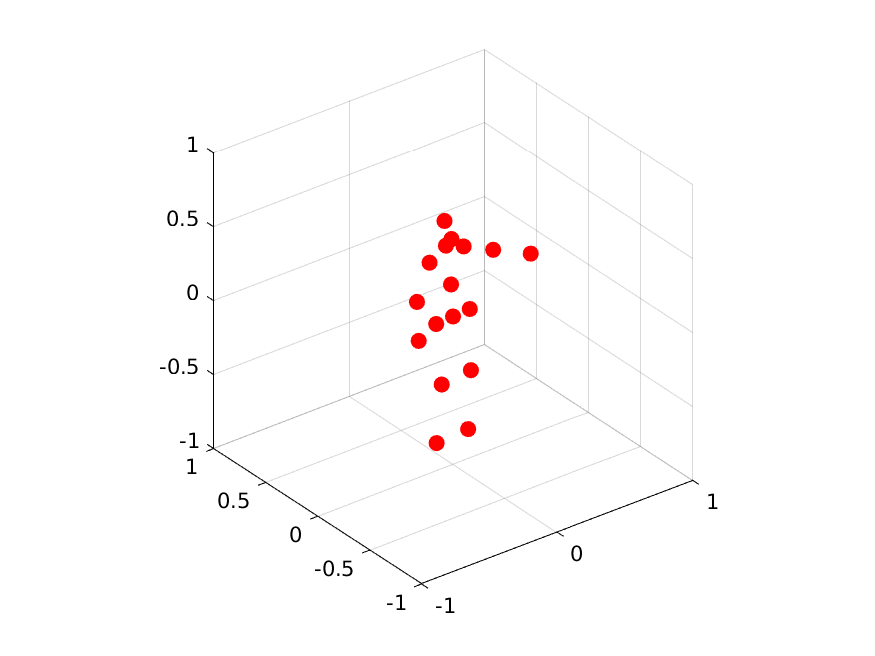}& 
   
         \includegraphics[width=0.32\linewidth]{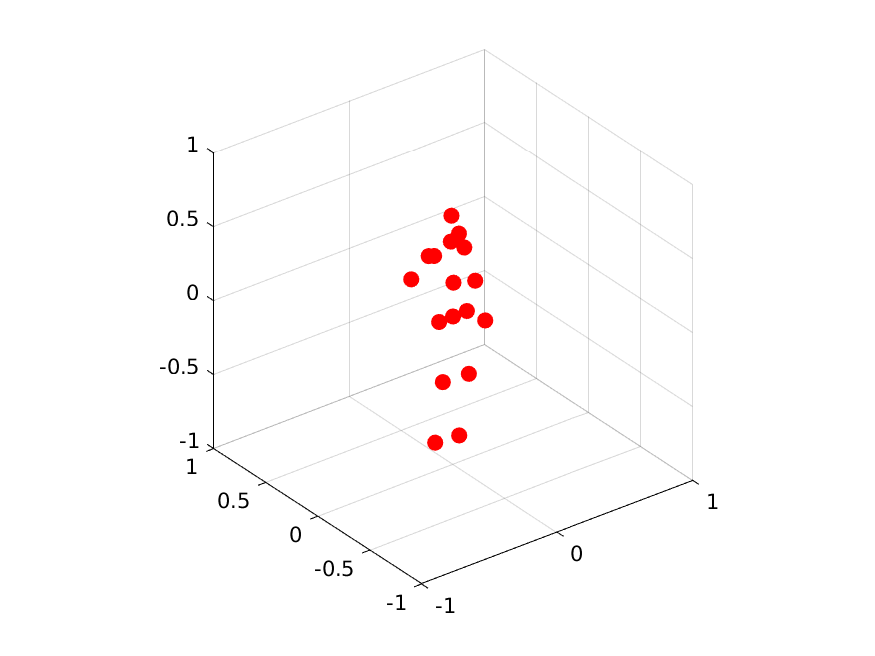}&

         \includegraphics[width=0.32\linewidth]{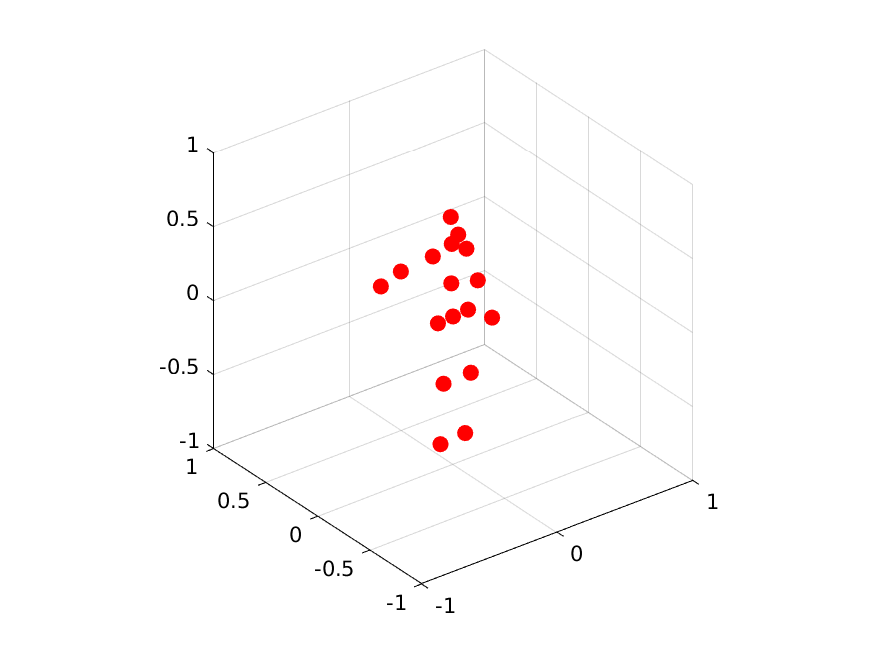}\\
         
         \raisebox{2.0\normalbaselineskip}[0pt][0pt]{\rotatebox[origin=c]{90}{mmMesh~\cite{mmMesh}}} & 
         
         \includegraphics[width=0.32\linewidth]{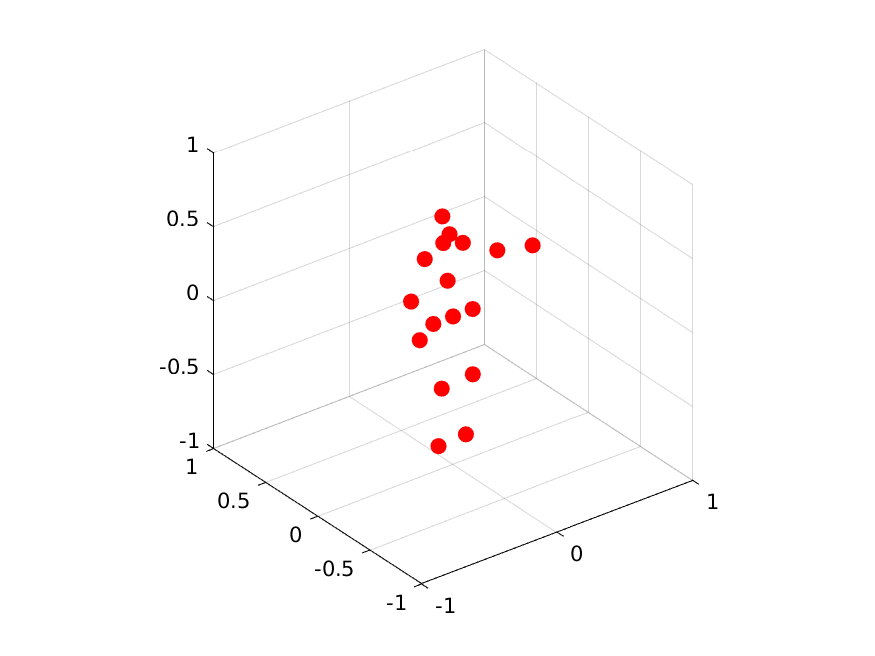} &
         \includegraphics[width=0.32\linewidth]{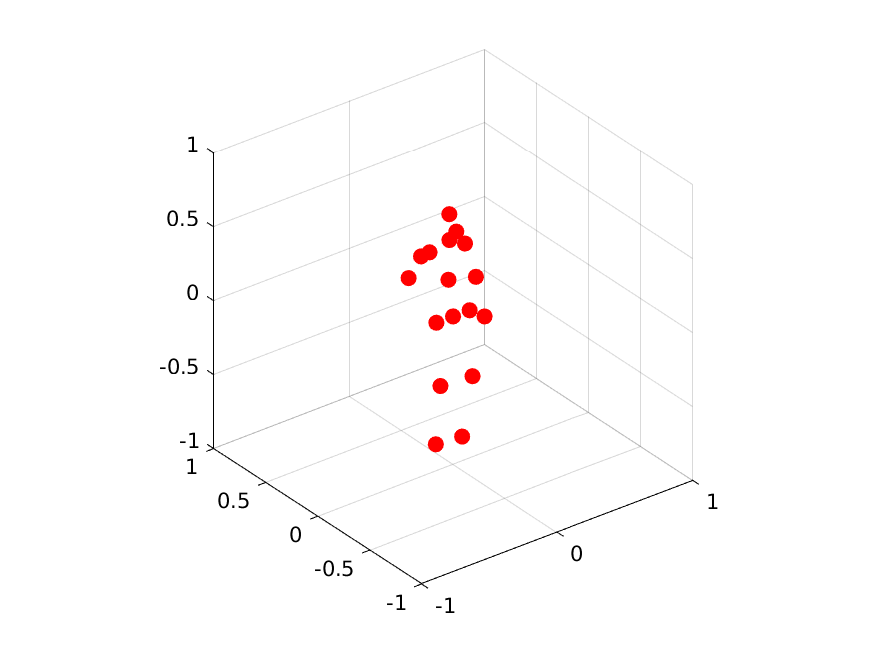} &

         \includegraphics[width=0.32\linewidth]{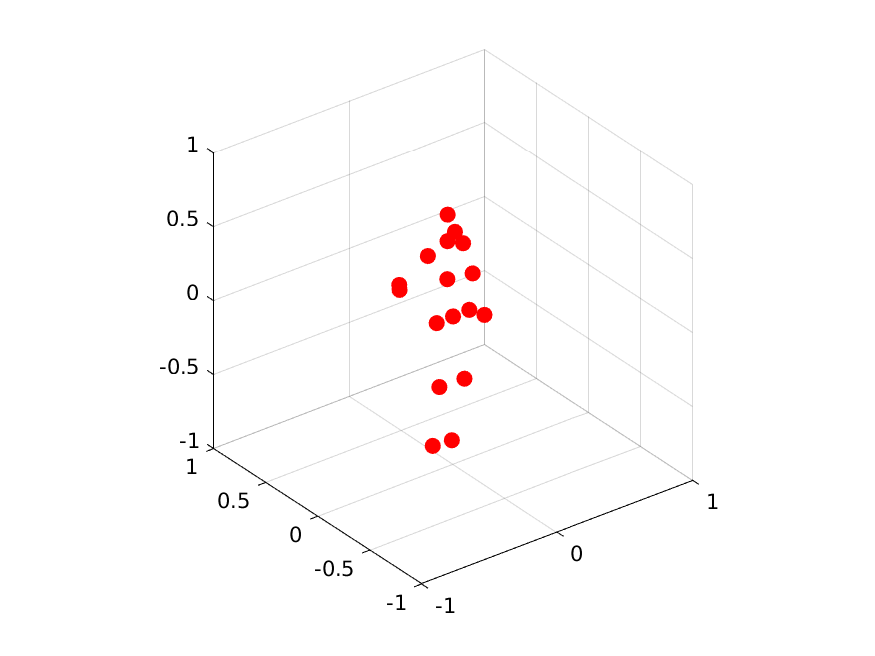} \\
         
         \raisebox{2.0\normalbaselineskip}[0pt][0pt]{\rotatebox[origin=c]{90}{Ours}} & 

         \includegraphics[width=0.32\linewidth]{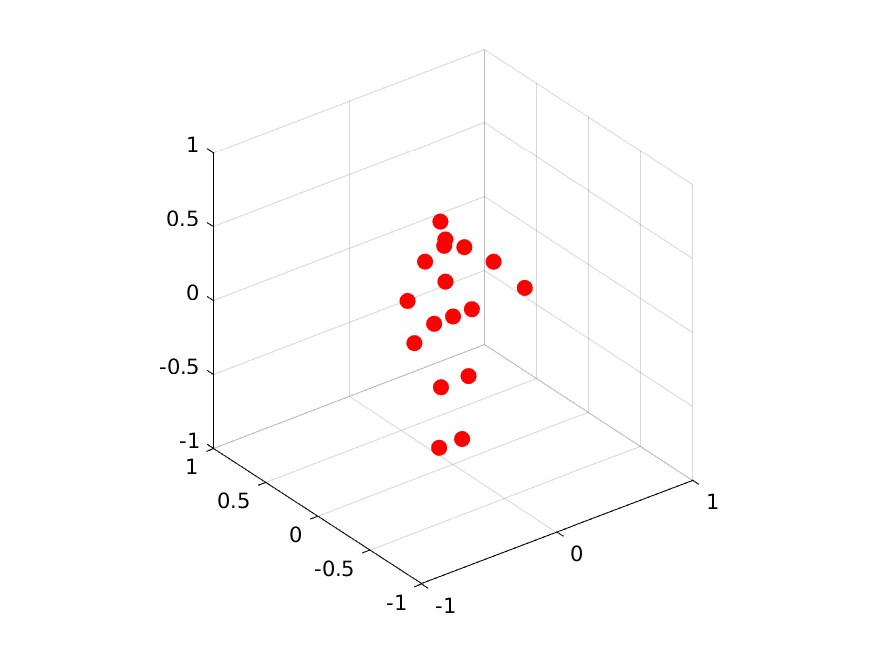} &

         \includegraphics[width=0.32\linewidth]{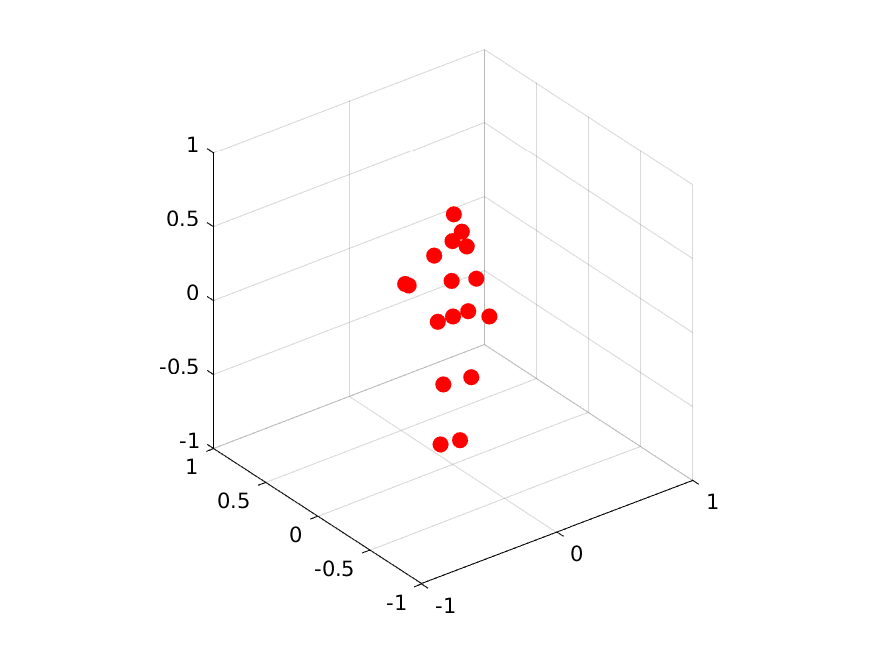} &

         \includegraphics[width=0.32\linewidth]{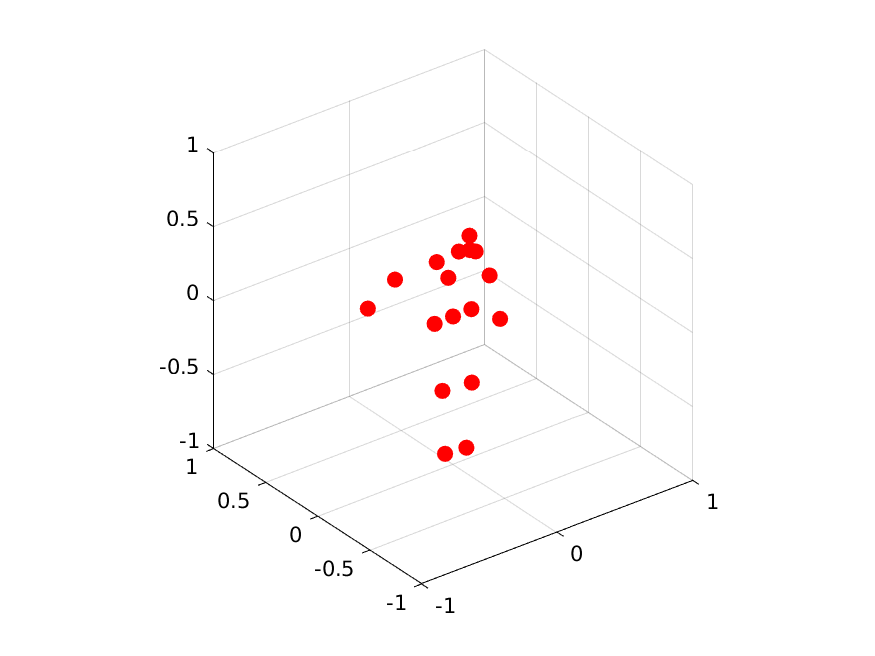} \\

    \end{tabular}
    \vspace{-0.8em}
    \caption{Comparison of 3D qualitative results. GT: Ground-truth keypoints generated by VideoPose3D~\cite{videopose3d}. Ours: The results generated by VideoPose3D based on our 2D predicted results.}
    \label{fig:comp_3d}
\end{figure}
\begin{figure}[t]
    \centering
    \setlength\tabcolsep{0.2em}
    \begin{tabular}{cccc}
        \includegraphics[width=0.22\linewidth]{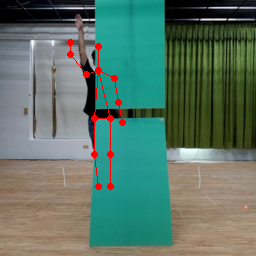} & 
        \includegraphics[width=0.22\linewidth]{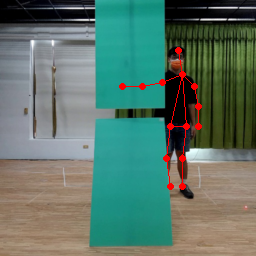} &
        \includegraphics[width=0.22\linewidth]{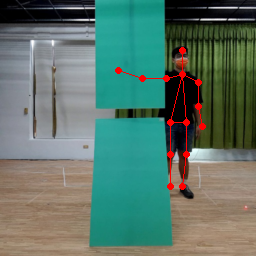} &
        \includegraphics[width=0.22\linewidth]{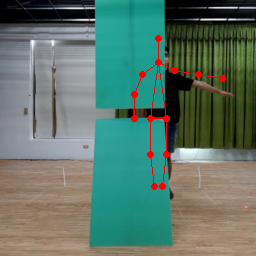} \\
        \includegraphics[width=0.22\linewidth]{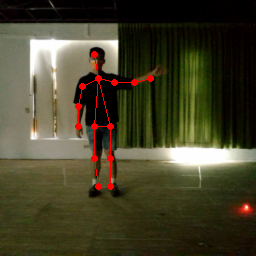} &
        \includegraphics[width=0.22\linewidth]{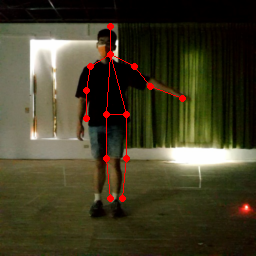} &
        \includegraphics[width=0.22\linewidth]{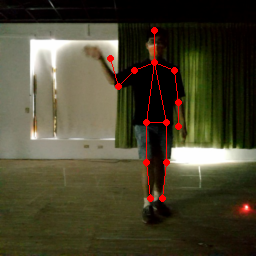} &
        \includegraphics[width=0.22\linewidth]{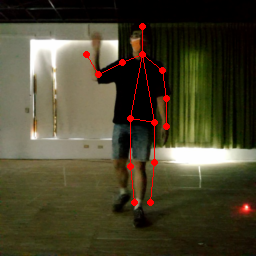} \\
        \includegraphics[width=0.22\linewidth]{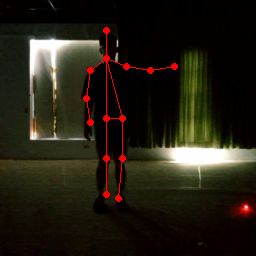} & \includegraphics[width=0.22\linewidth]{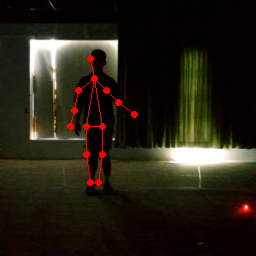} &
        \includegraphics[width=0.22\linewidth]{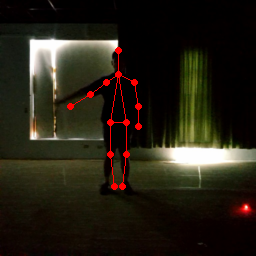} &
        \includegraphics[width=0.22\linewidth]{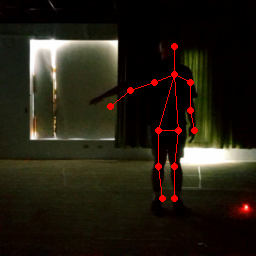} \\
    \end{tabular}
    \vspace{-0.8em}
    \caption{Our results under different conditions. Top row: obstacles in front of the person. Middle row: low-light condition. Bottom row: dark condition.}
    \label{fig:test_results}
\end{figure}

\subsection{Qualitative Results}
Fig.~\ref{fig:comp_qualitative} and Fig.~\ref{fig:comp_3d} demonstrate the 2D and 3D qualitative results of different types of actions. Our predicted keypoints mostly follow the ground-truths.
It is worth noting that the ground-truth keypoints from HRNet~\cite{hrnet} precisely follow the actions in the RGB frames, demonstrating a quality level similar to manual annotations. 
We also test our method on an extra test set under conditions of obstacle, low light, and no light (Fig.~\ref{fig:test_results}). Our proposed method produces promising results under these conditions, showing that radar signals and our method are robust to adverse lighting and visual blocking conditions. 


\section{Conclusion}
The paper introduces a novel human pose estimation scheme for mmWave radar sensing inputs, develops a radar pre-processing method, and proposes CSAM for multi-scale radar feature fusion along with PRGCN for pose confidence refinement. Our major findings and limitations include (1) the additionally extracted velocity information is critical to HPE; (2) our proposed CSAM and PRGCN improve the performance of 2D and 3D HPE as compared to the state-of-the-art RF-based and pointcloud-based methods; (3) the predicted results by taking radar signals as the input are qualitatively robust to dark, low-light, and visually blocked conditions; (4) the low azimuth and elevation angle resolution due to the limited number of antennas on the radar module~\cite{TI} result in the ambiguity between some actions, e.g., waving hands and raising hands.
Lastly, the potential of this line of research is very promising. Further investigation, e.g., extending our system to multi-person 2D, 3D HPE with more complex poses, is highly encouraged.

{\small
\bibliographystyle{ieee_fullname}
\bibliography{main}
}

\end{document}